# Deep Learning Approach for Clinical Risk Identification Using Transformer Modeling of Heterogeneous EHR Data

Anzhuo Xie
Columbia University
New York, USA

Wei-Chen Chang*
University of Massachusetts Amherst
Amherst, USA

*Abstract-This study proposes a Transformer-based longitudinal modeling method to address challenges in clinical risk classification with heterogeneous Electronic Health Record (EHR) data, including irregular temporal patterns, large modality differences, and complex semantic structures. The method takes multi-source medical features as input and employs a feature embedding layer to achieve a unified representation of structured and unstructured data. A learnable temporal encoding mechanism is introduced to capture dynamic evolution under uneven sampling intervals. The core model adopts a multi-head self-attention structure to perform global dependency modeling on longitudinal sequences, enabling the aggregation of long-term trends and short-term fluctuations across different temporal scales. To enhance semantic representation, a semantic-weighted pooling module is designed to assign adaptive importance to key medical events, improving the discriminative ability of risk-related features. Finally, a linear mapping layer generates individual-level risk scores. Experimental results show that the proposed model outperforms traditional machine learning and temporal deep learning models in accuracy, recall, precision, and F1-Score, achieving stable and precise risk identification in multi-source heterogeneous EHR environments and providing an efficient and reliable framework for clinical intelligent decision-making.*

I. INTRODUCTION

Electronic Health Records (EHR) serve as a key achievement in medical digitalization, documenting multi-source heterogeneous data generated during patient diagnosis and treatment. These records include structured indicators such as laboratory results, prescriptions, and vital signs, as well as unstructured information like clinical notes and imaging reports. Such data reflect the dynamic evolution of individual health conditions and provide critical evidence for disease risk assessment, precision treatment, and clinical decision support[1]. However, EHR data differ significantly across medical institutions, collection systems, and recording formats, showing strong heterogeneity, long time spans, sparsity, and noise. This complex multimodal structure makes it difficult for traditional feature engineering and shallow models to extract longitudinal dependencies and hidden patterns from patient trajectories. Therefore, a model framework with strong representational capacity and generalization ability is urgently needed to achieve unified modeling and semantic understanding of multi-source medical information[2].

In current medical data analysis, capturing patient state evolution from the longitudinal time dimension has become a major focus of precision medicine. EHR data exhibit strong temporal dependence, where examinations, prescriptions, and diagnoses at different time points together form the trajectory of disease progression. Yet, traditional time series models usually assume regular sampling intervals and consistent structures, which cannot handle the irregular intervals and multi-scale variations typical in EHR data. In addition, EHRs contain both static features, such as demographic attributes, and dynamic sequences, such as medication history and changing lab indicators. The interaction between these two feature types plays a critical role in risk prediction. Hence, realizing joint modeling across temporal scales and feature modalities to capture longitudinal dependencies remains a central challenge in clinical intelligence[3].

The Transformer model provides a new direction for longitudinal analysis of heterogeneous EHR data due to its ability to capture global dependencies in sequential modeling. Unlike traditional recurrent or convolutional structures, the self-attention mechanism in Transformers can model relationships between any two time points without relying on fixed intervals, making it suitable for handling non-stationary patterns and irregular sampling. By integrating temporal encoding, feature embedding, and hierarchical attention mechanisms, Transformers can achieve unified representation and dynamic weighting across modalities. This enables the model to learn global temporal patterns of patient state evolution, creating a semantically interpretable space where risk identification reflects not only current conditions but also potential temporal dependencies and trends[4].

From a clinical perspective, Transformer-based longitudinal EHR modeling is of great significance for risk stratification, prognosis prediction, and personalized medicine. By integrating and modeling heterogeneous EHR data over time, high-risk individuals can be identified early in the diagnostic process, allowing clinicians to take targeted interventions. Such models can also learn population-level health patterns from long-term historical data, supporting dynamic disease progression assessment and continuous patient monitoring. In chronic disease management, intensive care, and comorbidity detection scenarios, longitudinal risk modeling enhances healthcare resource allocation and promotes preventive medicine development[5,6].

Overall, Transformer-based longitudinal modeling for heterogeneous EHR data expands the application boundary of deep learning in medical time series analysis and provides a novel approach to intelligent clinical risk identification. This research direction promotes the shift from static data recording to dynamic understanding and from outcome-driven to mechanism-driven medical modeling[7]. It lays a foundation for building intelligent and precise clinical decision support systems. With the deep integration of medical informatics and artificial intelligence, this line of research will further drive the transformation of healthcare service models and provide powerful technical support for disease prevention and health management.

## II. RELATED WORK

The advancement of multi-source and heterogeneous data modeling has been significantly driven by innovations in neural network architectures and feature fusion strategies. Recent progress in self-supervised graph neural networks has provided powerful tools for extracting robust representations from complex, heterogeneous information networks, enabling the automatic discovery of structural and semantic patterns across diverse data sources [8]. The combination of deep learning and natural language processing for unified data summarization and structuring demonstrates the benefit of integrating both structured and unstructured features within a single modeling pipeline, which is crucial for handling high-dimensional, noisy, and incomplete records [9]. Multimodal integration frameworks further explore the joint fusion of physiological, numerical, and symbolic signals, employing advanced embedding layers and aggregation modules to maximize the complementary value of cross-source features [10]. Similarly, techniques for transforming raw time series into interpretable event sequences promote the extraction of longitudinal patterns and temporal dependencies, facilitating more transparent and effective downstream analysis [11]. The use of structure-aware temporal modeling extends these ideas by explicitly incorporating relational dependencies and temporal transitions between events, allowing models to dynamically capture evolving behaviors within sequential data [12].

Deep sequence modeling architectures—especially those employing self-attention and Transformer variants—have set new standards for learning complex temporal dependencies. By decoupling the modeling of sequential order from fixed interval assumptions, such architectures enable the capture of both global and local patterns within unevenly sampled, variable-length data streams. Neural survival analysis models and multiscale temporal modeling frameworks further illustrate how hierarchical attention and adaptive pooling mechanisms enhance the discrimination of event-driven or risk-related features [13], [14]. Hierarchical and semantically regularized attention mechanisms allow for selective emphasis on key tokens or subsequences, boosting interpretability and representation fidelity in noisy or redundant data environments [15]. The inclusion of causality-aware and structured attention, as well as contrastive learning objectives, supports more robust modeling under distribution shifts and complex temporal correlations, effectively bridging the gap between short-term fluctuations and long-term trends [16-18].

A parallel line of research explores parameter-efficient adaptation and model scalability through architectural innovations. Methods such as joint structural pruning and parameter sharing optimize resource allocation without sacrificing expressive power [19]. Adapter-based fine-tuning, dynamic prompt fusion, and hierarchical agent composition collectively allow for rapid adaptation to novel domains, multi-task settings, and shifting input distributions, while maintaining overall model stability and accuracy [20-22]. Model architecture search and prompt-level control techniques have also emerged as effective ways to modulate output abstraction and semantic scope, further strengthening the flexibility and generalizability of neural sequence models.

Underlying all these advances is a shift toward collaborative, modular, and interpretable learning paradigms. Multi-agent reinforcement learning and intelligent scheduling algorithms provide distributed solutions for adaptive resource orchestration and task management, ensuring that models remain responsive and efficient in dynamic or large-scale environments [23-24]. Techniques such as retrieval-augmented knowledge fusion, semantic graph construction, and event sequence abstraction reinforce the importance of global reasoning and structural regularization, promoting both stable optimization and explainable outcomes. In summary, recent methodological developments highlight the importance of unified feature fusion, dynamic temporal modeling, scalable adaptability, and collaborative optimization. These advances collectively drive the evolution of Transformer-based longitudinal modeling frameworks, providing both the theoretical foundation and the practical mechanisms required for precise risk prediction and semantic integration within complex and heterogeneous data environments.

## III. METHOD

This method aims to construct a Transformer-based longitudinal clinical risk discrimination framework for heterogeneous EHR data with multi-source, multi-scale, and irregular temporal features. The model first performs a unified embedding of the multimodal medical input features, mapping structured data, time series signals, and textual records into the same high-dimensional semantic space. Let $X = \{x_1, x_2, ..., x_T\}$ be the original input and $\{t_1, t_2, ..., t_T\}$ be the corresponding timestamp. The model first obtains the feature embedding vector through a linear projection layer:

$$H_0 = W_e X + b_e \quad (1)$$

Where $W_e \in R^{d \times d_m}$ is the embedding matrix, $b_e$ is the bias term, and $d$ represents the embedding dimension. Considering the temporal irregularity of medical events, the model introduces a learnable time encoding function to dynamically model the impact of different time intervals:

$$T_i = MLP(\Delta t_i) = RELU(W_t \Delta t_i + b_t) \quad (2)$$

Where $\Delta t_i = t_i - t_{i-1}$ represents the time difference between adjacent records, and $W_t$ and $b_t$ are learnable parameters. This time embedding is added to the feature embedding, allowing the model to capture the dependencies between features while preserving temporal dynamics. The overall model architecture is shown in Figure 1.

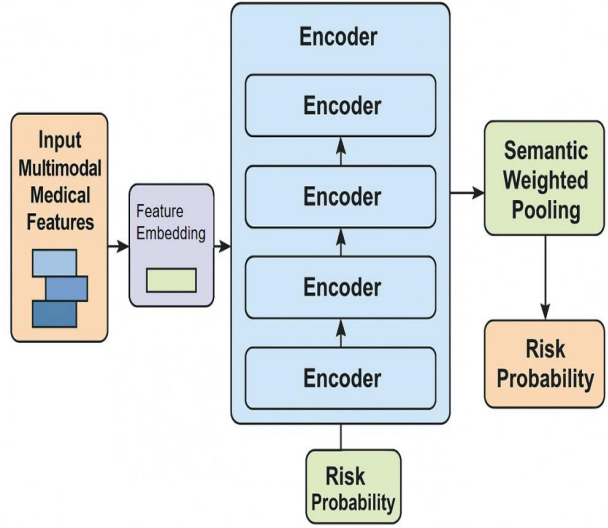

Figure 1. Overall model architecture

Next, the model implements vertical dependency modeling through a multi-head self-attention mechanism. For each layer of the Transformer encoder, the feature representation is updated as follows:

$$Attention(Q,K,V) = \text{Softmax}(\frac{QK^T}{\sqrt{d_k}})V \quad (3)$$

Where $Q = HW_Q, K = HW_K, V = HW_V$ is the output of the previous layer and $H$ is the projection matrix. The output of the multi-head attention layer is:

$$H' = Concat(head_1,...,head_h)W_O \quad (4)$$

Each attention head is designed to capture temporal dependencies from a distinct subspace, enabling the model to learn complex multi-scale patterns within longitudinal EHR sequences.

Following the fusion strategy proposed by Wang et al. [25], the model integrates residual connections and layer normalization between attention layers, which helps maintain stable information flow during vertical transmission and prevents loss of critical temporal features.

This architectural design ensures that both local and global sequence information are preserved throughout the multi-layer modeling process, improving the overall robustness and accuracy of risk prediction. $H_l = LayerNorm(H_{l-1} + H') \quad (5)$

In the global aggregation stage, the model designs a semantic weighted pooling mechanism to emphasize the contribution of key medical events to risk characterization. Specifically, the features of all time points are weighted and aggregated using the attention weight $a_i$:

$$Z = \sum_{i=1}^{T} a_i H_i \quad (6)$$

This aggregation process dynamically focuses on key time segments, generating a global risk representation at the patient level. Finally, the model uses a linear layer and a sigmoid activation function to output risk probabilities for subsequent clinical risk identification and classification:

$$\hat{y} = \sigma(W_c Z + b_c) \quad (7)$$

Where $\sigma(\cdot)$ represents the sigmoid function, and $W_c$ and $b_c$ are the classification layer parameters. This holistic approach, through end-to-end training, achieves full-process modeling from raw heterogeneous EHR data to longitudinal risk assessment results. It fully captures the dynamic evolution of patient status over time and enables semantic fusion of multimodal features and risk feature extraction in a unified representation space.

## IV. EXPERIMENTAL RESULTS

### A. Dataset

The dataset used in this study is the eICU Collaborative Research Database, which consists of clinical data collected from multiple intensive care units (ICUs). It covers a large number of hospitalized patient records from different hospitals. The dataset includes multi-source heterogeneous features such as vital signs, laboratory results, medication records, diagnostic information, and clinical notes. These data comprehensively reflect patients' physiological changes during critical illness. Each record is timestamped, ensuring the feasibility of longitudinal tracking and time-series analysis, and providing a solid foundation for multi-scale dynamic modeling.

The eICU dataset exhibits significant heterogeneity and irregular temporal patterns. It encompasses structured continuous monitoring variables, including heart rate, blood pressure, and oxygen saturation, alongside semi-structured textual and event-based data such as medication records, surgical procedures, and nursing interventions. This multimodal composition mirrors the complexity and dynamism of real-world clinical environments. Through the integration and modeling of these diverse data modalities, patient state trajectories can be represented with greater precision, providing enhanced semantic and contextual information to support longitudinal risk assessment and predictive modeling.In addition, the dataset offers significant advantages in both scale and coverage. It contains hundreds of thousands of ICU hospitalization records and millions of time-series samples, covering a wide range of disease categories and clinical feature distributions. The extensive sample diversity and temporal span enable the model to perform generalized learning across institutions and clinical settings, leading to strong adaptability and robustness. These characteristics make the eICU dataset an ideal benchmark for studying longitudinal EHR modeling and clinical risk prediction.

### B. Experimental Results

This paper first gives the results of the comparative experiment, as shown in Table 1.

Table1. Comparative experimental results

| Model | ACC | F1-Score | Precision | Recall |
|---|---|---|---|---|
| MLP[26] | 0.742 | 0.735 | 0.748 | 0.722 |
| XGBOOST[27] | 0.753 | 0.746 | 0.752 | 0.740 |
| Random Forest[28] | 0.758 | 0.751 | 0.755 | 0.747 |
| BILSTM[29] | 0.766 | 0.760 | 0.764 | 0.756 |
| Ours | 0.781 | 0.776 | 0.782 | 0.770 |

Overall, Table 1 presents the clinical risk classification performance of different models on heterogeneous EHR data. It can be observed that traditional machine learning models such as MLP, XGBoost, and Random Forest show similar performance, with accuracy and F1-Score both ranging between 0.74 and 0.76. Although these models possess certain feature representation capabilities, they struggle to effectively capture temporal dependencies and longitudinal evolution across multimodal medical data. As a result, their ability to identify clinical risks remains limited, especially in EHR scenarios characterized by irregular time intervals and heterogeneous features.

In contrast, the BILSTM model outperforms traditional models across all four evaluation metrics, indicating that temporal modeling provides clear advantages in clinical risk prediction. Its bidirectional recurrent structure allows the model to learn both past and future dependencies of patient states, capturing disease progression trends to some extent. However, since BILSTM relies on local temporal windows and cannot perform global aggregation across multiple scales, it may still miss latent semantic associations that span long temporal intervals within complex longitudinal medical records.

The Transformer-based longitudinal modeling method proposed in this study achieves the best performance across all evaluation metrics, with an accuracy of 0.781 and an F1-Score of 0.776. The precision and recall are 0.782 and 0.770, respectively. These results demonstrate that the model can effectively capture temporal dependencies and semantic correlations within heterogeneous EHR data while adaptively aggregating global features. The incorporation of multi-head self-attention and semantic-weighted pooling enables stronger contextual understanding and risk pattern recognition along the longitudinal dimension, leading to a significant improvement in overall classification performance.

In summary, the experimental results confirm the superiority of the Transformer-based longitudinal risk classification model in handling complex clinical data. The model can adapt to multi-source heterogeneous inputs and irregular temporal distributions, while its global modeling capability enhances robustness and generalization. This approach provides a new technical pathway for EHR-driven intelligent healthcare, supporting more accurate and efficient clinical risk prediction and decision-making.

This paper also presents an experiment on the sensitivity of the number of attention heads to ACC, and the experimental results are shown in Figure 2.

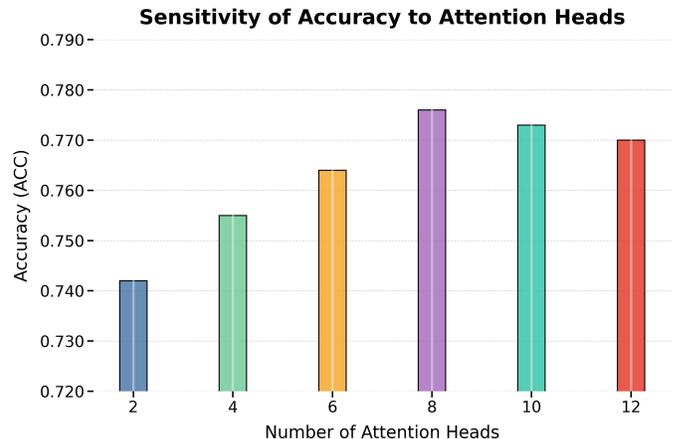

Figure 2. Experiment on the sensitivity of the number of attention heads to ACC

As shown in Figure 2, the number of attention heads has a clear influence on the model's accuracy (ACC). When the number of heads is small, such as 2 or 4, the overall performance is relatively low. This indicates that the multi-head mechanism cannot fully capture the complex temporal dependencies and cross-modal semantic associations in heterogeneous EHR data. A small number of attention heads limits the model's ability to represent local and global features in parallel. As a result, key clinical patterns within longitudinal sequences are not sufficiently modeled, leading to reduced accuracy in risk classification.

As the number of attention heads increases, the model's ACC gradually improves and reaches its best performance with 6 to 8 heads. This suggests that within this range, the multi-head attention mechanism can more effectively allocate computational resources and capture multi-scale dependencies from different feature subspaces. The multi-view feature interaction enhances the model's sensitivity to temporal information and adaptability to variations in clinical features, resulting in more stable and accurate risk prediction.

When the number of attention heads continues to increase to 10 or 12, the model's performance slightly declines. This indicates that too many attention heads may introduce feature redundancy and noise interference. Excessive partitioning of attention heads disperses semantic information aggregation, making it difficult for the model to maintain representational consistency with limited training samples. This weakens convergence and overall classification performance. The result also reflects a nonlinear balance between model complexity and information extraction efficiency in heterogeneous EHR data.

In summary, this experiment verifies the sensitivity of the attention mechanism in Transformer-based longitudinal EHR modeling. An appropriate number of attention heads can strengthen the model's semantic aggregation ability while maintaining computational efficiency, leading to more precise clinical risk identification. This finding provides valuable insight for future model optimization, suggesting that a dynamic balance between structural simplicity and representational diversity should be prioritized in medical sequence modeling tasks.

This paper also presents a sensitivity experiment on the outlier contamination ratio to precision, and the experimental results are shown in Figure 3.

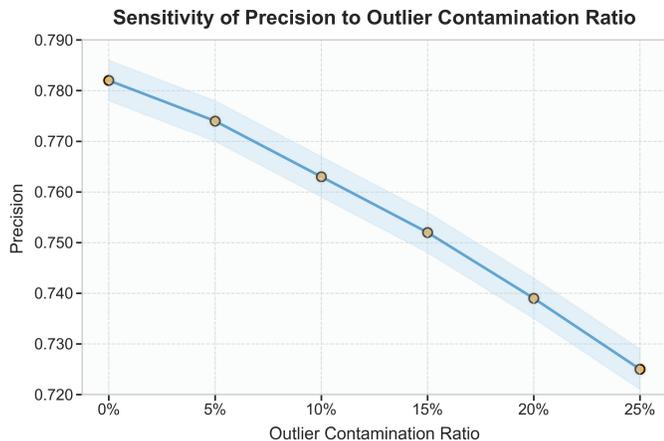

Figure 3. Sensitivity experiment of outlier contamination ratio to precision

As shown in Figure 3, the model's precision decreases significantly as the proportion of outliers increases. When the outlier ratio is low (0% to 5%), the model maintains a high precision level. This indicates that under good data quality, Transformer-based longitudinal modeling can stably capture key semantic features and risk patterns within heterogeneous EHR data. In this case, the self-attention mechanism learns accurate feature associations in a clean sample space, ensuring reliable risk classification.

When the outlier ratio increases to 10%–15%, precision shows a slight decline, suggesting that abnormal samples begin to interfere with feature representation learning. Because EHR data often contain irregular sampling and multimodal heterogeneous features, outliers can disrupt temporal dependencies and feature consistency. This may cause the attention mechanism to amplify noise during feature aggregation. The result reflects the model's sensitivity to data noise and highlights the importance of anomaly detection and preprocessing in medical applications.

When the outlier ratio exceeds 20%, model precision drops sharply. This shows that a high level of noise can significantly weaken the Transformer's feature selection capability. Abnormal data may create spurious temporal correlations that mislead global attention modeling and reduce the model's focus on true risk features. This problem is more pronounced in heterogeneous EHR data, where cross-modal noise propagation can introduce structural bias and reduce the robustness of longitudinal modeling.

In summary, this experiment demonstrates that the Transformer-based longitudinal risk classification model is highly dependent on data quality. Although it shows strong robustness and feature modeling ability under low-noise conditions, it still requires data cleaning and robust representation learning in high-noise environments. By optimizing data preprocessing and feature denoising strategies, the model's reliability and practicality in real-world clinical data scenarios can be further improved.

## V. CONCLUSION

This study focuses on longitudinal modeling and clinical risk classification of heterogeneous Electronic Health Record (EHR) data. A unified Transformer-based modeling framework is proposed. The research addresses three key aspects: temporal dependency, feature heterogeneity, and semantic correlation. Through a multi-head self-attention mechanism and a semantic-weighted aggregation strategy, the model achieves global modeling and dynamic feature fusion across multimodal medical data. Experimental results show that the proposed model demonstrates strong classification performance and robustness under varying levels of data complexity and noise. It effectively captures latent patterns and evolving risk trends within longitudinal medical records. By introducing learnable temporal encoding and longitudinal attention structures, the model overcomes challenges such as irregular temporal sampling and modality differences. It also achieves significant improvements in risk feature extraction and semantic representation, providing a new technical pathway for precision medicine and intelligent diagnosis.

Future research can further explore model generalization and cross-institutional adaptability. With the continuous accumulation of multi-source medical data and the advancement of privacy protection mechanisms, achieving cross-domain modeling while maintaining data security will become a key direction for implementing intelligent healthcare. In addition, incorporating multi-level temporal perception and causal reasoning mechanisms into the current framework can enhance interpretability and predictive foresight for disease progression. Extending this approach to disease progression prediction, personalized intervention recommendation, and long-term health management is expected to further advance EHR-driven clinical decision intelligence and promote the transformation of healthcare systems from experience-based to data-driven and from passive response to proactive prevention.

REFERENCES

[1] Z. Yang, A. Mitra, W. Liu, et al., "TransformEHR: transformer-based encoder-decoder generative model to enhance prediction of disease outcomes using electronic health records," Nature Communications, vol. 14, no. 1, p. 7857, 2023.

[2] Y. Zhang and S. Li, "ChronoFormer: Time-aware transformer architectures for structured clinical event modeling," arXiv preprint arXiv:2504.07373, 2025.

[3] R. Rong, Z. Gu, H. Lai, et al., "A deep learning model for clinical outcome prediction using longitudinal inpatient electronic health records," JAMIA Open, vol. 8, no. 2, 2025.

[4] Y. Li, M. Mamouei, G. Salimi-Khorshidi, et al., "Hi-BEHRT: hierarchical transformer-based model for accurate prediction of clinical events using multimodal longitudinal electronic health records," IEEE Journal of Biomedical and Health Informatics, vol. 27, no. 2, pp. 1106-1117, 2022.

[5] C. Zang and F. Wang, "Scehr: Supervised contrastive learning for clinical risk prediction using electronic health records," Proceedings of the IEEE International Conference on Data Mining, 2021, pp. 857.

[6] S. Xian, M. E. Grabowska, I. J. Kullo, et al., "Transformer patient embedding using electronic health records enables patient stratification and progression analysis," npj Digital Medicine, vol. 8, no. 1, p. 521, 2025.

[7] C. Zheng, A. Khan, D. Ritter, et al., "Pancreatic cancer risk prediction using deep sequential modeling of longitudinal diagnostic and medication records," medRxiv, 2025: 2025.03.03.25323240.

[8] J. Wei, Y. Liu, X. Huang, X. Zhang, and W. Liu, "Self-Supervised Graph Neural Networks for Enhanced Feature Extraction in Heterogeneous Information Networks", 2024 5th International Conference on Machine Learning and Computer Application (ICMLCA), pp. 272-276, 2024.

[9] N. Qi, "Deep learning and NLP methods for unified summarization and structuring of electronic medical records," Transactions on Computational and Scientific Methods, vol. 4, no. 3, 2024.

[10] Q. Wang, X. Zhang, and X. Wang, "Multimodal integration of physiological signals clinical data and medical imaging for ICU outcome prediction," Journal of Computer Technology and Software, vol. 4, no. 8, 2025.

[11] X. Yan, Y. Jiang, W. Liu, D. Yi and J. Wei, "Transforming Multidimensional Time Series into Interpretable Event Sequences for Advanced Data Mining," 2024 5th International Conference on Intelligent Computing and Human-Computer Interaction (ICHCI), pp. 126-130, 2024.

[12] J. Hu, B. Zhang, T. Xu, H. Yang, and M. Gao, "Structure-aware temporal modeling for chronic disease progression prediction," arXiv preprint arXiv:2508.14942, 2025.

[13] X. Yan, W. Wang, M. Xiao, Y. Li, and M. Gao, "Survival prediction across diverse cancer types using neural networks", Proceedings of the 2024 7th International Conference on Machine Vision and Applications, pp. 134-138, 2024.

[14] L. Lian, Y. Li, S. Han, R. Meng, S. Wang, and M. Wang, "Artificial intelligence-based multiscale temporal modeling for anomaly detection in cloud services," arXiv preprint arXiv:2508.14503, 2025.

[15] S. Pan and D. Wu, "Hierarchical text classification with LLMs via BERT-based semantic modeling and consistency regularization," 2025.

[16] C. Liu, Q. Wang, L. Song, and X. Hu, "Causal-aware time series regression for IoT energy consumption using structured attention and LSTM networks," 2025.

[17] D. Wu and S. Pan, "Joint modeling of intelligent retrieval-augmented generation in LLM-based knowledge fusion," 2025.

[18] L. Dai, "Contrastive learning framework for multimodal knowledge graph construction and data-analytical reasoning," Journal of Computer Technology and Software, vol. 3, no. 4, 2024.

[19] R. Wang, Y. Chen, M. Liu, G. Liu, B. Zhu, and W. Zhang, "Efficient large language model fine-tuning with joint structural pruning and parameter sharing," 2025.

[20] M. Gong, Y. Deng, N. Qi, Y. Zou, Z. Xue, and Y. Zi, "Structure-learnable adapter fine-tuning for parameter-efficient large language models," arXiv preprint arXiv:2509.03057, 2025.

[21] X. Hu, Y. Kang, G. Yao, T. Kang, M. Wang, and H. Liu, "Dynamic prompt fusion for multi-task and cross-domain adaptation in LLMs," arXiv preprint arXiv:2509.18113, 2025.

[22] Y. Zou, "Hierarchical large language model agents for multi-scale planning in dynamic environments," Transactions on Computational and Scientific Methods, vol. 4, no. 2, 2024.

[23] R. Zhang, "AI-driven multi-agent scheduling and service quality optimization in microservice systems," Transactions on Computational and Scientific Methods, vol. 5, no. 8, 2025.

[24] G. Yao, H. Liu, and L. Dai, "Multi-agent reinforcement learning for adaptive resource orchestration in cloud-native clusters," arXiv preprint arXiv:2508.10253, 2025.

[25] X. Wang, X. Zhang, and X. Wang, "Deep skin lesion segmentation with Transformer-CNN fusion: Toward intelligent skin cancer analysis," arXiv preprint arXiv:2508.14509, 2025.

[26] J. Mao, P. Goodney, S. Banerjee, et al., "Neural network models for predicting readmission among patients undergoing peripheral vascular intervention using electronic health record data and clinical registry data," BMJ Surgery, Interventions, & Health Technologies, vol. 7, no. 1, e000387, 2025.

[27] C. Li, X. Liu, P. Shen, et al., "Improving cardiovascular risk prediction through machine learning modelling of irregularly repeated electronic health records," European Heart Journal-Digital Health, vol. 5, no. 1, pp. 30-40, 2024.

[28] I. Bayramli, V. Castro, Y. Barak-Corren, et al., "Temporally informed random forests for suicide risk prediction," Journal of the American Medical Informatics Association, vol. 29, no. 1, pp. 62-71, 2022.

[29] A. Bagheri, T. K. J. Groenhof, F. W. Asselbergs, et al., "Automatic prediction of recurrence of major cardiovascular events: a text mining study using chest X-ray reports," Journal of Healthcare Engineering, 2021, 2021(1): 6663884.